\documentclass{article}
\usepackage{spconf,amsmath,graphicx,hyperref}

\usepackage{times}
\usepackage{epsfig}
\usepackage{graphicx}
\usepackage{amsmath}
\usepackage{amssymb}
\usepackage[ruled,linesnumbered]{algorithm2e}
\usepackage{multirow}
\usepackage{makecell}
\usepackage{balance}
\usepackage{float}
\usepackage{multirow}
\usepackage{subfigure}
\usepackage{subcaption}
\usepackage{booktabs}
\usepackage{setspace}

\title{Supervised Makeup Transfer with a Curated Dataset: Decoupling Identity and Makeup Features for Enhanced Transformation}
\name{$^{1}$Qihe Pan
        \qquad $^{2}$Yiming Wu
        \qquad $^{1}$Xing Zhao$^{\star}$
        \qquad $^{1}$Liang Xie
        \qquad $^{1}$Guodao Sun
        \qquad $^{1}$Ronghua Liang
}
  
  \address{$^{1}$School of Computer Science and Technology, Zhejiang University of Technology, Zhejiang, China \\
      $^{2}$ The University of Hong Kong, Hong Kong, China}

\begin{document}
%
\maketitle
%


\begin{abstract}
Diffusion models have recently shown strong progress in generative tasks, offering a more stable alternative to GAN-based approaches for makeup transfer. Existing methods often suffer from limited datasets, poor disentanglement between identity and makeup features, and weak controllability. To address these issues, we make three contributions. First, we construct a curated high-quality dataset using a train–generate–filter–retrain strategy that combines synthetic, realistic, and filtered samples to improve diversity and fidelity. Second, we design a diffusion-based framework that disentangles identity and makeup features, ensuring facial structure and skin tone are preserved while applying accurate and diverse cosmetic styles. Third, we propose a text-guided mechanism that allows fine-grained and region-specific control, enabling users to modify eyes, lips, or face makeup with natural language prompts. Experiments on benchmarks and real-world scenarios demonstrate improvements in fidelity, identity preservation, and flexibility. Examples of our dataset can be found at: \href{https://makeup-adapter.github.io/}{https://makeup-adapter.github.io/}.
\end{abstract}

\section{Introduction}

Makeup transfer aims to apply reference makeup styles to a source face while preserving the person’s original identity. As an important task in image editing, it is widely applied in virtual try-on, beauty enhancement, and digital entertainment. The core difficulty lies in generating realistic results that faithfully reflect reference makeup while maintaining source identity, a balance that remains challenging in practice.

Early studies primarily adopted generative adversarial networks (GANs) \cite{li2018beautygan,gu2019ladn,jiang2020psgan,chang2018pairedcyclegan,yan2022beautyrec}, which achieved visually convincing results but suffered from training instability and limited controllability. More recently, diffusion models \cite{ho2020denoising,song2020denoising,rombach2022high,ramesh2022hierarchical,zhang2023adding,gal2022image,zhang2024stable,jin2024toward,sun2025shmt} have demonstrated stronger generative capacity and finer editing flexibility, making them promising for makeup transfer. Despite these advances, two major challenges remain.  
\begin{figure}[h]
  \centering
  \begin{minipage}[b]{0.8\linewidth}
    \centering
    \includegraphics[width=0.8\linewidth]{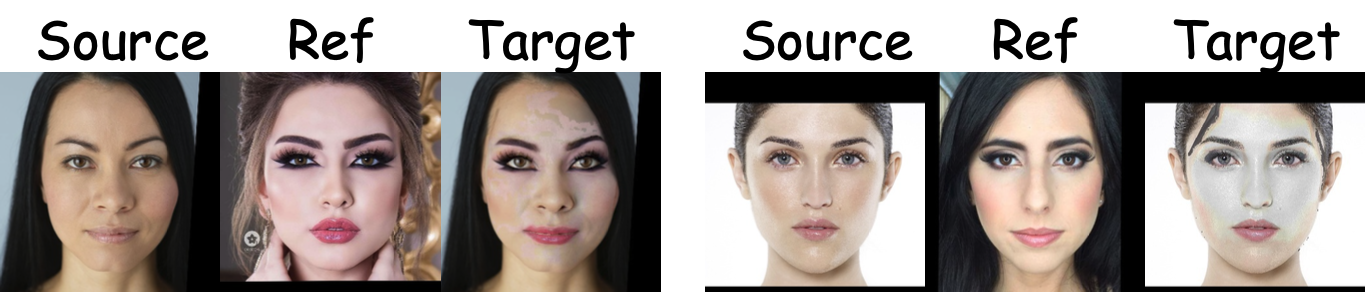}\\
    (a) Bad cases in training dataset LADN.
  \end{minipage}

  \begin{minipage}[b]{0.8\linewidth}
    \centering
    \includegraphics[width=0.8\linewidth]{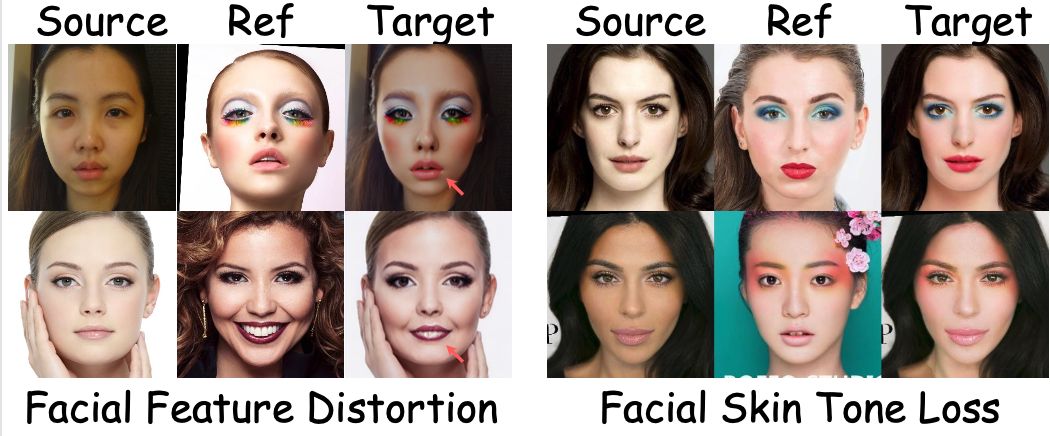}\\
    (b) Failure cases generated by existing method.
  \end{minipage}

  \vspace{-0.1cm}
  \caption{Limitations of existing datasets and methods.}
  \label{fig:limitations}
\end{figure}
The first challenge is disentangling identity and makeup features. In practice, identity-specific information—such as facial geometry and skin tone—often becomes entangled with makeup attributes like colors and textures. Without proper separation, models tend to either alter the original identity or fail to reproduce reference makeup details. This trade-off between identity preservation and makeup fidelity persists across both GAN- and diffusion-based methods.  

The second challenge concerns the scarcity of high-quality paired datasets. Ideally, supervised training requires before-and-after images of the same individual, yet such pairs are extremely rare. Existing works resort to pseudo-paired data generated by histogram matching or geometric distortion \cite{sun2022ssat,yang2022elegant}, which fail to capture realistic spatial and stylistic patterns. Furthermore, widely used datasets such as LADN \cite{gu2019ladn}, MT \cite{li2018beautygan}, and Wild-MT \cite{jiang2020psgan} contain numerous low-quality samples, as illustrated in Fig.~\ref{fig:limitations}, limiting their usefulness for robust training.  

In this work, we address these issues from both model and data perspectives. On the modeling side, we propose a diffusion-based framework with explicit disentanglement of identity and makeup features, achieved via dedicated feature extractors and a cycle-reconstruction mechanism. This design enables faithful makeup transfer while maintaining strong identity consistency. On the data side, we introduce a novel \textit{train–generate–filter–retrain} strategy to construct a high-quality paired dataset. Starting from an initial model, we generate candidate pairs, filter them for visual quality and consistency, and retrain with the refined data. This iterative pipeline substantially improves dataset fidelity and diversity, supporting more effective supervised training. Our contributions are summarized as follows:  
\begin{itemize}
    \item We design a diffusion-based framework that explicitly disentangles identity and makeup features, effectively resolving the trade-off between identity preservation and makeup fidelity. 
    \item We construct a high-quality paired dataset through a novel \textit{train–generate–filter–retrain} pipeline, which progressively expands, filters, and refines data to enable robust supervised training.  
    \item We validate our approach through extensive experiments and user studies, showing that the proposed method and curated dataset set a new benchmark in makeup transfer.  
\end{itemize}

By jointly advancing model design and dataset construction, our work provides a comprehensive solution to long-standing challenges in makeup transfer and opens new opportunities for practical applications.

\section{Related Work}

\subsection{Makeup Transfer}

Early approaches used facial landmarks and warping \cite{li2015simulating}, which were efficient but unrealistic. With deep generative models, GAN- and diffusion-based methods became mainstream. BeautyGAN \cite{li2018beautygan} applied a dual-GAN with histogram matching, PSGAN \cite{jiang2020psgan} introduced attention, and others \cite{sun2022ssat, sun2024content, yang2022elegant, gu2019ladn, nguyen2021lipstick} explored local discriminators or masks. Diffusion-based methods brought more stability and control: Stable-Makeup \cite{zhang2024stable} generated paired data via LEDITS \cite{tsaban2023ledits}, TinyBeauty \cite{jin2024toward} proposed lightweight designs, and SHMT \cite{sun2025shmt} adopted self-supervised training. Beyond 2D, works such as Geneavatar \cite{bao2024geneavatar}, BareSkinNet \cite{umetani2022bareskinnet}, and \cite{yang2023makeup,yang2024makeup} explore 3D priors for better geometric consistency, though at higher cost.  

On the dataset side, Stable-Makeup showed that LEDITS++ can synthesize paired makeup/non-makeup images, inspiring our design. BeautyBank \cite{lu2025beautybank} provides a larger collection but with limited quality. In contrast, we develop a train–generate–filter–retrain pipeline that refines data into a high-fidelity paired dataset for robust, identity-preserving transfer.

\section{Methods}

\textbf{Model Architecture:}  
Our framework consists of three components: an identity feature extractor, a makeup feature extractor, and a U-Net generator. The identity branch encodes structural attributes such as facial geometry and skin tone, while the makeup branch captures cosmetic attributes like colors and textures. These representations, together with text embeddings, are fused in the U-Net via Mixed-guided-attention to enable controllable, fine-grained transfer.

\textbf{Notation:}  
We denote an image as $I_iM_j$, where $i$ specifies the identity and $j$ the makeup style. When $j=0$, $I_iM_0$ corresponds to a bare-faced input. This notation is used consistently throughout the paper.

\textbf{Data Pre-processing:}  
Faces are segmented into semantic regions (face, eyes, lips) with masks. During training, text prompts are aligned with these regions to guide localized transfer, preventing style leakage and ensuring accurate region-specific application.

\begin{figure}[htb]
  \centering
  \includegraphics[width=1.0\linewidth]{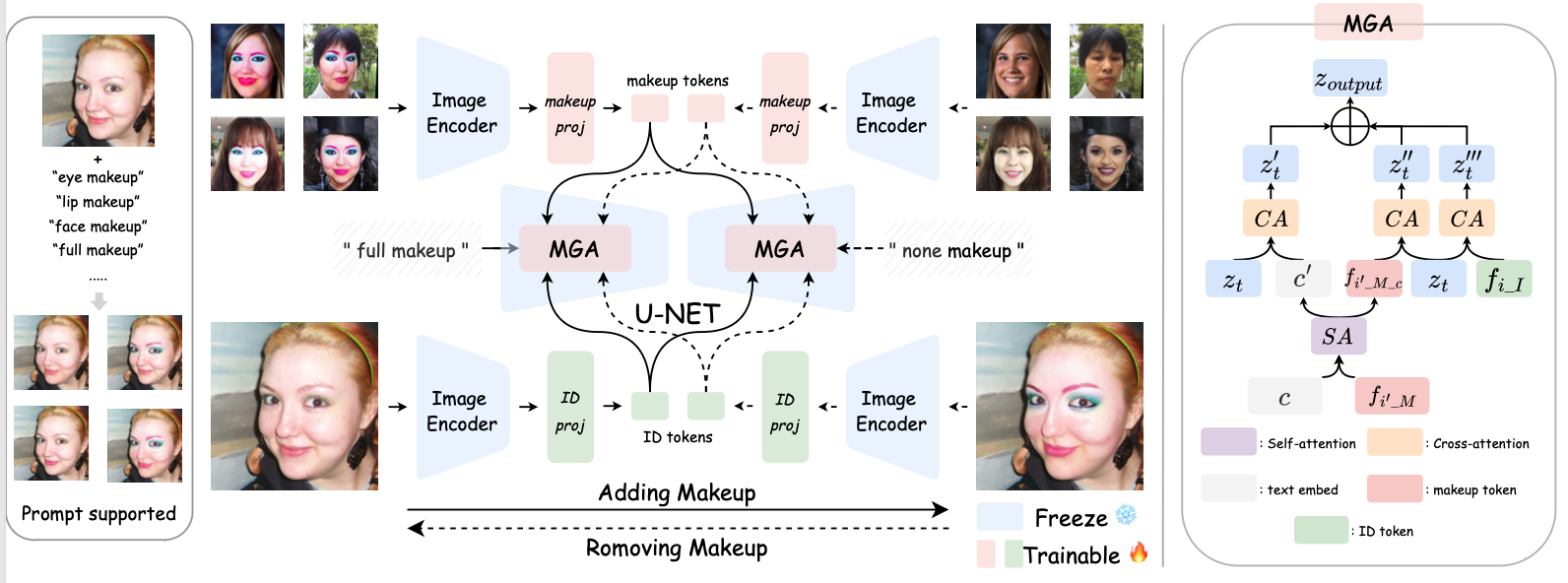}
    \caption{
    Framework overview. 
    \textbf{Middle:} training pipeline with disentanglement and reconstruction. 
    \textbf{Left:} prompt-based training, where region-specific prompts (eyes, face, lips...) map to target images.
    \textbf{Right:} Mixed-Guided Attention (MGA) combining text, identity, and makeup features. 
    }
  \label{fig:trainingpipeline}
\end{figure}

\subsection{ID-Makeup Decouple}

A key challenge in makeup transfer is separating identity from cosmetic style. To address this, we adopt a dual-branch encoding scheme. Given an input $I_iM_j$, the encoder $\mathcal{E}(\cdot)$ extracts features $f_i$, which are projected into two embeddings: identity $f_{i_I}$ and makeup $f_{i_M}$, through projectors $P_I$ and $P_M$. For another image $I_{i'}M_{j'}$, we similarly obtain $f_{i'_I}$ and $f_{i'_M}$. The generator $\mathcal{G}$ then combines $f_{i_I}$ with $f_{i'_M}$ to reconstruct $\hat{I}_iM_{j'}$, optimized by the reconstruction loss:
\[
\mathcal{L}_{Makeup} = \| \mathcal{G}(f_{i_I}, f_{i'_M}) - I_iM_{j'} \|_2.
\]

To further enforce disentanglement, we require the recovery of bare-faced images. Pairing $f_{i_I}$ with a non-makeup embedding $f_{i'_M}(M=0)$ guides the generator to reconstruct $I_iM_0$, ensuring that identity embeddings preserve geometry and skin tone while makeup embeddings remain style-specific:
\[
\mathcal{L}_{ID} = \| \mathcal{G}(f_{i_I}, f_{i'_M}(M=0)) - I_iM_0 \|_2.
\]

In addition, we adopt the denoising objective from Stable Diffusion, where noisy latents $z_t$ are conditioned on text $c$, identity features $f_{i_I}$, and makeup features $f_{i'_M}$. The model predicts the added noise $\epsilon$, minimizing:
\[
\mathcal{L}_{diffusion} = \mathbb{E}_{z_t,t,c,\epsilon \sim \mathcal{N}(0,1)} 
\| \epsilon - \epsilon_\theta(z_t,t,c,f_{i_I},f_{i'_M}) \|_2.
\]

The final objective combines all three terms:
\[
\mathcal{L}_{total} = \mathcal{L}_{diffusion} + \mathcal{L}_{Makeup} + \mathcal{L}_{ID}.
\]

This joint design provides both explicit reconstruction supervision and generative learning, enabling faithful identity preservation while flexibly applying diverse makeup styles.






\subsection{Mixed-Guided Attention}

To enable text-guided control, we propose Mixed-Guided Attention (MGA), which integrates identity features $f_{i_I}$, makeup features $f_{i_M}$, and text embeddings $c$ (from CLIP). First, self-attention is applied to $c$ and $f_{i_M}$, producing an updated token $f_{i_M}^c$ that combines reference style with textual description.  

Next, cross-attention is performed between the noisy latent $z_t$ and three sources: text $c$, the updated makeup token $f_{i_M}^c$, and the identity token $f_{i_I}$. This yields outputs $Z_1, Z_2, Z_3$, which are fused as:
\begin{equation}
Z_{\text{out}} = \lambda_{text} Z_1 + \lambda_{makeup} Z_2 + \lambda_{id} Z_3.
\end{equation}

The weights $\lambda$ provide explicit control over the trade-off between makeup strength and identity preservation. For example, setting $\lambda_{id}=0$ performs style-only transfer, while $\lambda_{makeup}=0$ prioritizes identity retention.  

By combining self-attention with decoupled cross-attention, MGA enables fine-grained, text-guided manipulation of cosmetic effects while maintaining structural consistency and realism.

\section{Dataset Construction}

To overcome the limitations of existing benchmarks, we propose a novel \textbf{train–generate–filter–retrain} strategy. This pipeline progressively integrates synthetic and realistic data, expands it through cross-identity generation, removes low-quality samples via filtering, and retrains the model with a curated dataset.

\subsection{Train: Initial Data Preparation}
\begin{figure}[H]
\begin{center}
\includegraphics[width=0.8\linewidth]{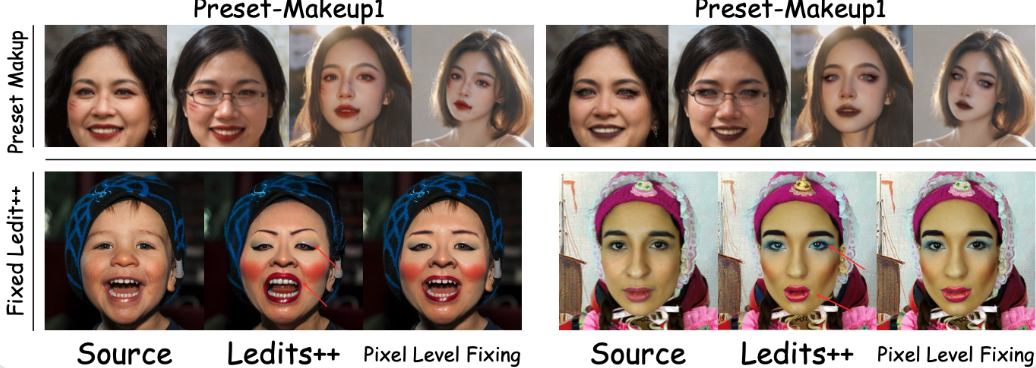}
\end{center}
\vspace{-0.5cm}
\caption{Examples of images using preset makeup and fixing ledit++ generated image.}
\label{fig:ledit_preset}
\end{figure}

We first construct a base dataset $\mathcal{D}_0$ from three sources: (i) synthetic makeup images generated with LEDITS++ on FFHQ faces, corrected by pixel-level warping for alignment ($\sim$50k pairs); (ii) publicly available makeup datasets; and (iii) commercial applications such as \textit{Meitu} and \textit{Xingtu}, where 100 preset styles were applied to 1,000 synthetic identities generated by StyleGAN2 and SDXL ($\sim$100k pairs). This yields a total of $\sim$150k paired samples $(I_iM_0, I_iM_j)$ for training the first-stage model $\mathcal{G}_1$ (see Fig.~\ref{fig:ledit_preset}).

\subsection{Generate–Filter–Retrain}
Given $\mathcal{G}_1$, we expand the dataset by cross-identity transfer.  
Let $\{I_aM_0\}_{a \in \mathcal{A}}$ be bare faces from identity set $\mathcal{A}$, and $\{I_bM_j\}_{b \in \mathcal{B}}$ be makeup faces from identity set $\mathcal{B}$ generated by LEDITS++. Stylized results are produced as
\begin{equation}
\hat{I}_aM_j = \mathcal{G}_1(f_{a_I}, f_{b_M}), \quad a \in \mathcal{A}, b \in \mathcal{B}.
\end{equation}

Since $\hat{I}_aM_j$ may contain artifacts, it is not used directly as a target. Instead, it serves as an auxiliary reference compared with the LEDITS++ image $I_bM_j$. Makeup similarity is computed using the projector $P_M$ and encoder $\mathcal{E}$:
\begin{equation}
\text{sim}(\hat{I}_aM_j, I_bM_j) =
\cos \big(P_M(\mathcal{E}(\hat{I}_aM_j)), P_M(\mathcal{E}(I_bM_j))\big).
\end{equation}

A filtering function is defined as:
\begin{equation}
\mathcal{F}(\hat{I}_aM_j, I_bM_j) =
\begin{cases}
1, & \text{if } \text{sim} \geq 0.7, \\
0, & \text{otherwise},
\end{cases}
\end{equation}
where the similarity threshold is set to $\tau = 0.7$. Only pairs with $\mathcal{F}=1$ are retained, yielding the curated dataset:
\begin{equation}
\mathcal{D}^\star = \{ (I_aM_0, I_bM_j) \mid \mathcal{F}(\hat{I}_aM_j, I_bM_j)=1 \}.
\end{equation}

After filtering, we obtain $\sim$150k clean samples. The final training set combines these with the original 50k LEDITS++ pairs, resulting in a curated dataset of about 200k pairs at $512 \times 512$ resolution (excluding the preset-style dataset from \textit{Meitu} and \textit{Xingtu}). The full pipeline can be summarized as:
\[
\mathcal{D}_0 \xrightarrow{\text{Train}} \mathcal{G}_1
\xrightarrow{\text{Generate}} \hat{\mathcal{D}}
\xrightarrow{\text{Filter}} \mathcal{D}^\star
\xrightarrow{\text{Retrain}} \mathcal{G}_2.
\]

\begin{figure}[H]
\begin{center}
\includegraphics[width=0.8\linewidth]{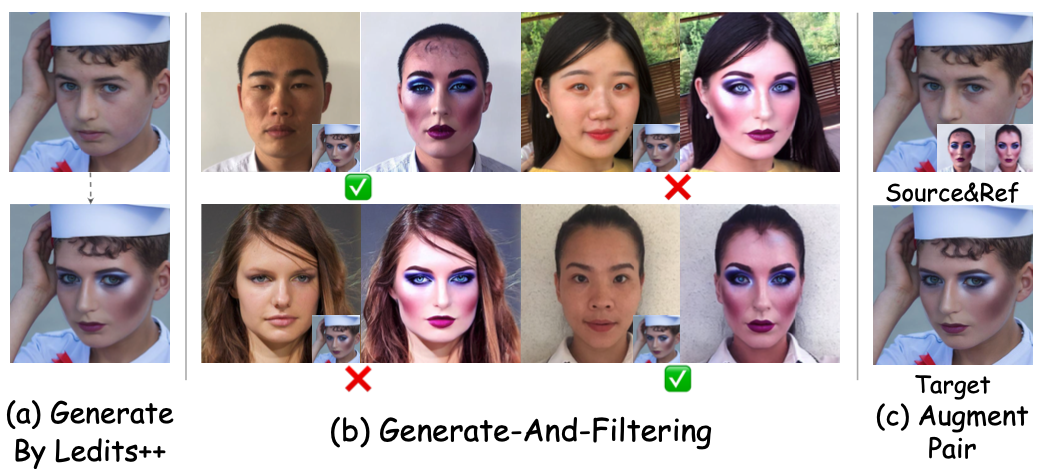}
\end{center}
\vspace{-0.5cm}
\caption{Examples of dataset augmentation using the generate–filter strategy, where generated results are compared with LEDITS++ references and low-similarity samples are discarded.}
\label{fig:gen3}
\end{figure}

\begin{table*}[t]
    \small
    \caption{Quantitative results of \emph{FID}, \emph{CLS} and \emph{Key-sim} on the MT, Wild-MT and LADN datasets and Our test dataset }
    \label{table:results}
    \centering
    \scalebox{0.6}{
        \begin{tabular}{l|c|c|c|c|c|c|c|c|c|c|c|c|c}
            \toprule
            \multirow{2}{*}{Methods} & \multicolumn{3}{c|}{MT}& \multicolumn{3}{c|}{Wild-MT} & \multicolumn{3}{c|}{LADN} & \multicolumn{4}{c}{Ours dataset} \\
            \cmidrule{2-4} \cmidrule{5-7} \cmidrule{8-10} \cmidrule{11-14}
            & FID   & CLS  & Key-sim & FID & CLS & Key-sim & FID & CLS & Key-sim & FID & CLS & Key-sim & FID-to-Real \\
            \midrule
            Stable-Makeup   &33.26& 0.682& 0.973& 64.64& 0.711& 0.968& 37.33& 0.767& 0.965 & 13.82 &0.622 & 0.985 & 6.75 \\
            \midrule
            SHMT-$h_0$ &32.24 &0.658 &0.976 & 51.54&0.668 &0.976 &38.97 &0.711 &0.978 & 20.84 & 0.577 & \textbf{0.992} & 12.01\\
                \midrule
            SHMT-$h_4$ &24.93 & \textbf{0.715}&0.953 &45.02  &\textbf{0.719}&0.954 &27.01 &\textbf{0.786} & 0.958 & 18.42 & 0.609 &0.989 & 11.03 \\
                \midrule
            Ours  & \textbf{ 12.07} &0.708 & \textbf{0.9889} & \textbf{17.86} & 0.712 &\textbf{ 0.9869} & \textbf{17.95} & 0.774 & \textbf{0.986} & \textbf{11.67} & \textbf{0.628} & \textbf{0.992} & \textbf{3.56} \\
            \bottomrule
    \end{tabular}}
\end{table*}

\section{Exprtiments}
\subsection{Experiment Settings}

\noindent\textbf{Datasets.} We train on our self-constructed dataset of 50K aligned image pairs (Section 4), without using MT or other public datasets. For evaluation, we follow HSMT and test on MT, Wild-MT, and LADN. Wild-MT contains large pose and expression variations, while LADN provides diverse and complex makeup styles, ensuring comprehensive evaluation.

\noindent\textbf{Training Details.} Our method is based on Stable-Diffusion-v1-5, trained at $512 \times 512$ resolution for 500k steps with a learning rate of $1\times10^{-6}$. We use 2 NVIDIA 4090 GPUs (batch size 8 per GPU), Adam optimizer, and set $\lambda_1=\lambda_2=1$ in the loss. Sampling is performed with DDIM (50 steps). The UNet backbone and its integration with our attention mechanism follow SD-v1.5.

\begin{table}[H]
\caption{User preference of quality and identity preservation.}
\vspace{-0.3cm}
\begin{center}
    \scalebox{0.5}{
    \begin{tabular}{>{\centering\arraybackslash}m{3cm}|>{\centering\arraybackslash}m{2.5cm}|>{\centering\arraybackslash}m{2cm}}
			\toprule
		      Methods    &  \emph{Stable-Makeup}  &  \emph{Ours} \\
		     \midrule
		      Makeup preference        &  35\% &  65\% \\
             \midrule
		      Identity preservation    & 18\% & 82\%  \\
			\bottomrule
    \end{tabular}}
\end{center}
\label{tab:userstudy}
\end{table}

\vspace{-0.7cm}

\begin{table}[H]
\caption{Ablation study of using different training dataset.}
\vspace{-0.3cm}
\centering
\scalebox{0.5}{
\begin{tabular}{c|c|c|c|c|c|c|c}
\toprule

\multicolumn{2}{c|}{Training Dataset}                 & \multicolumn{3}{c|}{LADN}   & \multicolumn{3}{c}{Our Dataset} \\ 
\midrule
Dataset 1+2 & Dataset 3  & FID   & CLS  & Key-sim & FID & CLS & Key-sim   \\ 
\midrule
\checkmark &                                & 22.38 & 0.736      & 0.943 & 14.85 & 0.596 & 0.970 \\
\midrule
\checkmark &\checkmark                      & 17.95 & 0.774      & 0.986 & 11.67 & 0.628 & 0.992 \\
\bottomrule
\end{tabular}}
\label{tab:ablation}
\end{table}
\vspace{-0.5cm}

\subsection{Comparison.}
\noindent\textbf{Baselines.} We compare with two diffusion-based state-of-the-art methods: Stable-Makeup\cite{zhang2024stable} and SHMT\cite{sun2025shmt}, using public code and pre-trained models. Our curated test set ensures fair evaluation, with results shown in Tab.~\ref{table:results}.

\noindent\textbf{Evaluation Metrics.} We use three metrics: FID (lower is better), CLS, and Key-sim (higher is better). Following SHMT, FID is computed between reference and transferred images, and on our dataset also with real makeup. CLS measures alignment with reference style, while Key-sim evaluates identity preservation, both using cosine similarity in DINO’s feature space.

\noindent\textbf{User Study.} To assess perceptual quality, we compared our method with Stable-Makeup at $512 \times 512$. Fifty participants evaluated 100 pairs on makeup fidelity, identity preservation, and overall quality. As shown in Tab.~\ref{tab:userstudy}, our approach was consistently preferred, especially for identity preservation and realism.

\noindent\textbf{Ablation Study.} We test different dataset combinations (Dataset 1, Dataset 2, Dataset 3). As shown in Tab.~\ref{tab:ablation}, Dataset 1+2 already gives solid results, but adding Dataset 3—integrating realistic, synthetic, and filtered data—further improves FID and CLS/Key-sim. This confirms our generate-and-filter strategy enhances data diversity and quality, leading to better fidelity, identity preservation, and robustness.

\vspace{-0.3cm}
\subsection{Text-Guided Region Makeup Transfer.}
Unlike mask-based methods, our approach enables text-guided region transfer without manual masks. Prompts such as “eye,” “lip,” or “full makeup” directly specify target regions, allowing precise control via natural language and avoiding mask limitations across face shapes.

\vspace{-0.3cm}
\subsection{Analysis.}
As shown in Fig.~\ref{table:results}, our method does not always achieve the best scores due to the trade-off between makeup fidelity and identity preservation, also observed in SHMT. Nevertheless, it maintains a strong balance, producing realistic and identity-consistent results. Remaining gaps are partly due to the limited quality and resolution of existing benchmarks.

\begin{figure}[h]
  \begin{center}
  \includegraphics[width=0.8\linewidth]{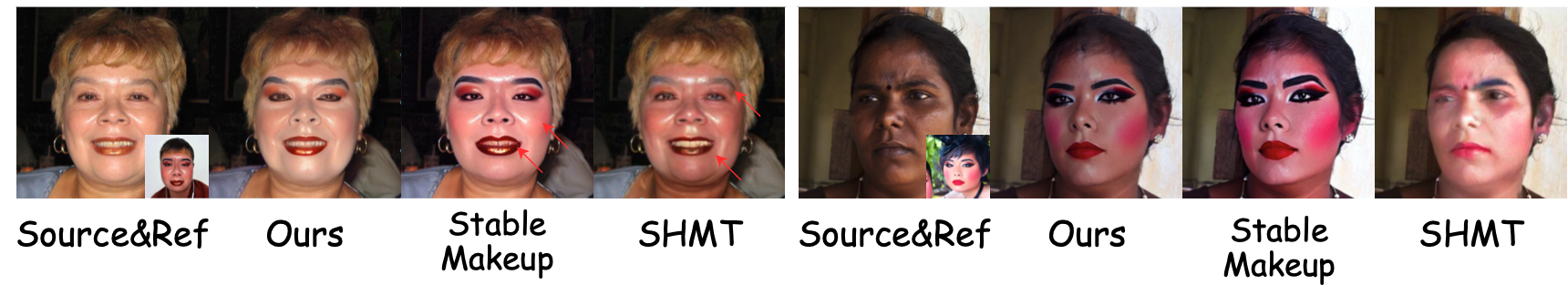}
  \end{center}
  \vspace{-0.5cm}
  \caption{visualization of comparing to other methods.}
  \label{fig:compare}
\end{figure}
\vspace{-0.5cm}

\begin{figure}[h]
  \begin{center}
  \includegraphics[width=0.8\linewidth]{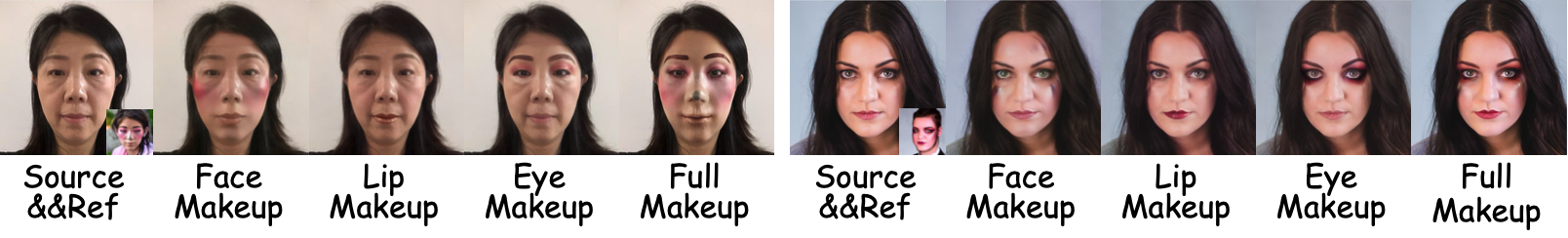}
  \end{center}

  \vspace{-0.5cm}
  
  \caption{visualization of fine control with text prompt.}
  \label{fig:textcond}
  
\end{figure}

\vspace{-0.5cm}
  
\vspace{-1em}

\section{Conclusion}
We propose a makeup transfer method that combines a curated dataset with Mixed-Guided Attention for identity-preserving transformation. Our high-resolution paired dataset offers greater style diversity and identity consistency than prior benchmarks. Future work will focus on improving robustness and realism in text-guided makeup transfer.
\vspace{-1em}

\section*{Acknowledgment}
This work was supported by the National Natural Science Foundation of China under Grant 62402441, Grant 62402440, Grant 62422607 and Grant 62432014;

\bibliographystyle{IEEEbib}
\bibliography{strings,refs}

\end{document}